# Theory of Acceleration of Decision Making by Correlated Time Sequences

Norihiro Okada,[1] Tomoki Yamagami,[2] Nicolas Chauvet,[2] Yusuke Ito,[1] Mikio Hasegawa,[1] and Makoto Naruse[2]

[1] Department of Electrical Engineering, Graduate School of Engineering, Tokyo University of Science, 6-3-1 Niijuku, Katsushika-ku, Tokyo 125-8585, Japan.

[2] Department of Information Physics and Computing, Graduate School of Information Science and Technology, The University of Tokyo, 7-3-1 Hongo, Bunkyo-ku, Tokyo 113-8656, Japan.

## Abstract

Photonic accelerators have been intensively studied to provide enhanced information processing capability to benefit from the unique attributes of physical processes. Recently, it has been reported that chaotically oscillating ultrafast time series from a laser, called laser chaos, provide the ability to solve multi-armed bandit (MAB) problems or decision-making problems at GHz order. Furthermore, it has been confirmed that the negatively correlated time-domain structure of laser chaos contributes to the acceleration of decision-making. However, the underlying mechanism of why decision-making is accelerated by correlated time series is unknown. In this study, we demonstrate a theoretical model to account for accelerating decision-making by correlated time sequence. We first confirm the effectiveness of the negative autocorrelation inherent in time series for solving two-armed bandit problems using Fourier transform surrogate methods. We propose a theoretical model that concerns the correlated time series subjected to the decision-making system and the internal status of the system therein in a unified manner, inspired by correlated random walks. We demonstrate that the performance derived analytically by the theory agrees well with the numerical simulations, which confirms the validity of the proposed model and leads to optimal system design. The present study paves the way for improving the effectiveness of correlated time series for decision-making, impacting artificial intelligence and other applications.

## 1. Introduction

Optics and photonics have been extensively studied for high-speed information processing in various applications, especially machine learning [1–5]. One of the important branches of the research frontier is reinforcement learning [6], wherein the impacts of photonics have been intensively examined [7–9]. The multi-armed bandit (MAB) problem regards decision-making in obtaining high rewards from multiple selections, called *arms*, wherein the best arm is initially unknown. MAB problems concern a difficult tradeoff known as the exploration-exploitation dilemma, which captures a fundamental aspect of reinforcement learning [6].

The physical properties of photons have been utilized in solving MAB problems [7,8]. In particular, chaotically oscillating ultrafast time series generated by semiconductor lasers, called laser chaos, have been successfully utilized in resolving two-armed bandit problems in GHz order, which we call laser chaos decision-maker hereafter [7]. As introduced below, the principle of the laser chaos decision-maker simply depends on the signal level comparison between the chaotically oscillating time series and the threshold level. It has also been demonstrated that such a level-comparison-based principle is scalable in a tree architecture, which can be experimentally demonstrated up to 64 arms [10].

Furthermore, the applications of laser chaos decision-makers have been studied to benefit from their prompt adaptation abilities in dynamically changing uncertain environments [11–14]. Takeuchi *et al.* applied laser chaos decision-making to channel selection problems in wireless communications [11], in which communication channels suffer from dynamically changing disturbances due to traffic, interference, or fading [15]. Kanemasa *et al.* extended the principle using laser chaos decision-maker to channel bonding in IEEE 802.11ac networks [12]. Further, Duan *et al.* optimized user-pairing in non-orthogonal multiple access (NOMA) systems by laser chaos decision-maker [13]. Moreover, Kanno *et al.* combined laser chaos-based decision-making with photonic reservoir computing, where adaptive model selection is realized to enhance the computing capability [14].

In [7], it was demonstrated that the autocorrelation inherent in laser chaos time series impacted the decision-making performances. Indeed, chaotic time series with negative maximum autocorrelation yield superior performances when compared with pseudorandom numbers, colored noise, and random shuffle surrogate data of the original laser chaos time series [7]. Furthermore, Okada *et al.* extensively examined the decision-making acceleration by laser chaos using surrogate analysis, such as the Fourier transform surrogate [16]. It was found that both statistical distributions of the amplitude of time series and negative autocorrelation therein impact decision-making performances [16].

In the literature, the usefulness of negative autocorrelation in time series has been theoretically analyzed regarding code division multiple access (CDMA) [17–19]. To achieve high performance in CDMA, the cross-correlation between the spreading sequences must be small. The optimal negative autocorrelation to minimize the interference has been mathematically derived and the chaotic map that generates the smallest cross-correlation was defined. In addition, ref. [19] clarifies that the negative autocorrelation that minimizes cross-correlation accelerates the performance of solution search algorithms for combinatorial optimization problems. An FIR filter to generate the optimal chaotic CDMA sequence was also proposed based on the negative autocorrelation analysis [20]. Moreover, the effectiveness of such optimal negative autocorrelation codes has been experimentally demonstrated using software-defined radio systems [21].

However, regarding decision-making, the fundamental underlying mechanism of how negative autocorrelation inherent in time series impacts the performance superiority is still unclear. That is, the results in the previous studies [7,16] are all limited in empirical findings. If the effectiveness of the negative autocorrelation in laser chaos or correlated time series for decision-making is theoretically grasped, it allows, for example, a systematic design approach to derive the optimal autocorrelation depending on given problem situations. Besides, the insights gained by mathematical modeling ensure the reliability of the effectiveness provided by the negative autocorrelation in time series.

In this study, we theoretically construct a model to account for the effect of negative autocorrelation in decision-making performances. The theory of the present study is inspired by correlated random walk [22,23]. Contrary to conventional random walks, which have transition probabilities independent of prior events, correlated random walks have probabilities dependent on prior events [22,23]. That is, the notion of correlated random walks allows us to represent state-dependent, different probability evolution dynamics. Such a theoretical architecture accounts for the interplay between the correlated time series and evolution of decision-making. We clarify the validity of the proposed theoretical model by confirming the excellent agreement of the decision-making performances derived analytically by the proposed model and by numerical simulations.

The rest of the paper consists of the following. Section 2 reviews the mechanism of laser chaos decision-maker. In Section 3, we introduce a numerical method to generate an arbitrary autocorrelation in time series, by which the relevance between autocorrelation and the resultant decision-making performance is systematically examined. Section 4, which is the most important contribution of this study, demonstrates the theoretical model of decision-making based on correlated time sequences. Section 5 demonstrates the agreement of the decision-making performances predicted by the proposed theory and numerical simulations. Section 6 concludes the paper.

## 2. Laser Chaos Decision Maker: Using Time Series for Decision Making

As mentioned in Section 1, the laser chaos time series allows ultrafast decision-making. Figure 1(a) schematically illustrates the architecture of the laser chaos decision-maker for a two-arm bandit problem, which is the scope of the present study [7]. The two arms are called slot machines A and B. Laser chaos is generated by subjecting a portion of the output light back to the laser by an externally arranged reflector, which is called delayed feedback. We compare the intensity level of the laser chaos with a certain threshold value, which is denoted by $T(t)$.

The decision-making is executed as follows. When the sampled value of the time series is above the threshold, the decision is to choose slot machine A; otherwise, slot machine B is selected. The threshold $T(t)$ is updated according to the result of the slot machine play. Overall, the threshold update is conducted under the assumption that the revised threshold will lead to the same decision in the subsequent decisions when the present action is successful, whereas the threshold is revised to the opposite direction when the present action is a failure [7,8,10].

More precisely, the values of threshold $T(t)$ are determined by
$$T(t) = k \times [TA(t)], \qquad (1)$$
Where $TA(t)$ is called the threshold adjuster and $[*]$ is the nearest integer to $*$. $[TA(t)]$ can take an integer value ranging from $-N$ to $N$, with $N$ being a natural number. Therefore, the number of levels that the threshold adjustor can take is $2N+1$. Here, $k$ is a coefficient to convert $[TA(t)]$ to $T(t)$.

$TA(t)$ is updated depending on the result of the action conducted at $t-1$:
$$\left.\begin{array}{ll} TA(t) = -\Delta + \alpha\, TA(t-1), & \text{if slot machine A wins} \\ TA(t) = +\Delta + \alpha\, TA(t-1), & \text{if slot machine B wins} \\ TA(t) = +\Omega + \alpha\, TA(t-1), & \text{if slot machine A fails} \\ TA(t) = -\Omega + \alpha\, TA(t-1), & \text{if slot machine B fails} \end{array}\right\}, \qquad (2)$$
where $\Delta$ denotes increment, which is given by $\Delta = 1$ in the present study. $\alpha$ is the forgetting parameter for weighting previous threshold adjuster variables, ranging from 0 to 1; i.e., $0 \leq \alpha \leq 1$. $\Omega$ is called the penalty parameter [7,8].

A hierarchical formation of such two-armed bandit problems has been proposed to deal with problems with more than two arms [10]. The elemental structure is the abovementioned two-armed situations with a dynamically updated threshold. This study focuses on two-arm situations as the first theoretical analysis on the laser chaos decision-maker. The analysis of cases with more than two arms can be done by extending the method proposed in this study; however, that will become a very complicated analysis. Therefore, we focus on a simple case in this study, and the cases with more than two arms will be our future work.

## 3. Effectiveness of Correlated Time Series on Decision Making

As described in Section 1, the performance of the two-armed bandit problem using laser chaos time series depends on the autocorrelation inherent therein [7,16]. The best performance is obtained when the autocorrelation of the time series exhibited its negative maximum [7]. Furthermore, the surrogate data analysis of laser chaos time series clarify the impact of time-domain correlation [10]. In this study, to examine the influence of correlations in time series in a systematic manner, we introduce an artistically constructed time-correlated time series and analyze its influence on decision-making performance.

We construct a time series whose amplitude follows a Gaussian distribution while having a determined autocorrelation by utilizing the Fourier transform surrogate method [24]. The various steps involved are as follows:

(1) Construct a time series $r(t)$ with $t$ ranging from 0 to $T-1$, where $T$ is the length of the time series. Here, we suppose that $r(t) = r(0)\,\lambda^t$. Specifically, $r(t) = \lambda\,r(t-1)$ holds, indicating that $r(t)$ undergoes a time correlation specified by $\lambda$ to its previous point $r(t-1)$. We call $\lambda$ the *autocorrelation coefficient* in this study.
(2) The Fourier transform of $r(t)$ is calculated and denoted by $R(f)$.
(3) The phase of $R(f)$ is revised by randomly assigned numbers, while the power spectrum is maintained. The revised Fourier-domain signal is denoted by $R'(f)$.
(4) By taking the inverse Fourier transform of $R'(f)$, a new time series is generated, which is denoted by $r'(t)$.

Through the process above, the autocorrelation of the resultant $r'(t)$ is equivalent to that of $r(t)$. However, the amplitude distribution of $r'(t)$ follows a Gaussian profile because of the randomized phase factors in the Fourier domain. The above-described process corresponds to a special case of Fourier transform surrogate [24].

Snapshots of the time series generated for the cases when the time correlation is specified by $\lambda = 0.8$, 0, and $-0.8$ are shown in Figures 1(b), 1(c), and 1(d), respectively. All of the time series signals appear random, but there are distinct differences in their autocorrelation. With $\lambda = 0.8$, the signal level at time $t$ is similar to the signals around that point; that is, radically large signal-level differences in consecutive data points are rarely observed (Figure 1(b)). Conversely, with $\lambda = -0.8$, meaning a strong negative autocorrelation, the signal at time $t$ has almost the exact opposite value to the surrounding data (Figure 1(d)). As a result, the time series exhibits a highly time-varying structure. Meanwhile, the histogram of the signal level of these time series follows the same Gaussian distribution.

It should be noted that the above-described Fourier transform surrogate-based procedure does not perfectly reproduce the experimentally observed laser chaos time series. This is because the correlation in the above process is determined only by $r(t) = \lambda\,r(t-1)$ in step (1), whereas the experimental laser chaos involves very long-range time correlations via delayed optical feedback. However, we consider that the Fourier transform surrogate-based method is quite beneficial for the present study for several reasons.

The first is that the correlation between two successive points can be specified by an arbitrary number, allowing $\lambda$ values smaller than even $-0.5$, which was experimentally not feasible, at least in the previous studies [7,10]. Therefore, systematic analysis is enabled for a wide range of $\lambda$. The second is that amplitude distributions are kept equivalent between each other even when $\lambda$ is configured to different values, which also allows us a clear examination of the impact of autocorrelation inherent in the time series.

For these reasons, we use the time series $r'(t)$ generated using the above process. We then analyze how the MAB performance depends on the autocorrelation specified by $\lambda$. In evaluating the performance of the MAB problem, we employ the *correct decision rate* (CDR). The $CDR(t)$ is defined as the ratio of selecting a slot machine with the highest reward probability at a time step $t$, and averaged over $m$ simulations or cycles. That is, $CDR(t)$ is expressed by

$$CDR(t) = \frac{1}{m} \sum_{i=1}^{m} C_i(t), \tag{3}$$

where $m$ is the number of cycles with different random initial conditions. Here, $C_i(t) = 1$ when the slot machine with the highest reward probability is selected at the $t$-th decision (or time $t$) of the $i$-th cycle. In other words, correct decision-making is conducted. Otherwise, $C_i(t) = 0$, meaning that correct decision-making is not executed. In the following simulations, $m = 60000$.

Figure 2 summarizes the calculated CDR at $t = 1000$ as a function of the autocorrelation coefficient $\lambda$ in several different reward environments and the setting of the decision-maker. The reward probability of the two slot machines, called machine A and machine B, are denoted by $P_A$ and $P_B$, respectively. For example, in Figure 2(a), $P_A$ and $P_B$ are given as 0.9 and 0.3, respectively. In this situation, the correct decision is to select machine A as it is the slot machine with the highest reward probability ($P_A > P_B$). In addition, the number of levels of threshold adjustor is five, and specified by $N = 2$. It should be emphasized that a higher CDR is obtained when the autocorrelation is negative; indeed, the best CDR is given by $\lambda = -0.6$.

Figures 2(b)–2(f) examine other reward settings and decision-maker conditions. Table 1 summarizes the reward probabilities of slot machines and the number of threshold levels $N$ for each MAB problem. In Figures 2(b) and 2(c), $P_A$ and $P_B$ are differently configured while maintaining the same threshold number as in Figure 2(a) (that is, $N = 2$). More specifically, the difference of $P_A$ and $P_B$ is only 0.1 in Figure 2(b) by setting $(P_A, P_B) = (0.6, 0.5)$. Similarly, the difference is 0.2 in Figure 2(c) by setting $(P_A, P_B) = (0.9, 0.7)$. That is, the difficulties in finding the best machine are configured differently. Here, it should be noted that the highest CDR is accomplished when the autocorrelation coefficient $\lambda$ is given by $-0.8$ and $-0.3$ in Figures 2(b) and 2(c), respectively. That is, the best decision-making is realized with negatively correlated time series.

The reward setting of $(P_A, P_B)$ in Figures 2(d), 2(e), and 2(f) is the same as in Figures 2(a), 2(b), and 2(c), respectively. The only difference is in the threshold value, which is specified by $N = 4$. The achieved CDR was different because of the change in the value of $N$. However, it should be noted that the highest CDR performances are all obtained with negative autocorrelation when $\lambda$ is given by $-0.6$, $-0.9$, and $-0.6$ in Figures 2(d), 2(e), and 2(f), respectively.

# 4. Theoretical Model of Decision Making Using Correlated Time Series

This section shows a mathematical model to account for the impact of correlated time series for decision-making. Here, we focus on two-armed bandit problems where two slot machines are called machines A and B. Figure 3 shows a conceptual architecture of the proposed model. We assume that slot machine A has a larger reward probability than slot machine B; that is, $P_A > P_B$. Therefore, the correct decision would be to choose slot machine A.

Here, we assume that the subjected time sequence takes either of the two signal levels specified by $+x$ or $-x$, which is denoted by sky blue marks in Figure 3. In the meantime, remember that the threshold level, $T(t)$ given by equation (1), takes in total $2N + 1$ different signal levels, each of which is represented by $-N, -N+1, \ldots, N-1, N$. Furthermore, we assume that the higher-level signal $+x$ satisfies $N - 1 < x < N$, meaning that the upper signal level of the incoming time series is below the maximum threshold level but greater than the second maximum threshold. Similarly, the lower signal level $(-x)$ satisfies $-N < -x < -N + 1$, indicating that the lower signal level of the subjected time series is above the minimum threshold level but less than the second minimum threshold.

Based on the decision-making principle described in Section 2, we summarize the decision-making process in the present situation. Let the signal level of the incoming time series at time $t$ and the threshold level at time $t$ be denoted by $s(t)$ and $T(t)$, respectively.

- If $T(t) = -N$, regardless of the signal level $s(t)$, slot machine A is selected. This is because $T(t) = -N < s(t)$ always holds since the minimum of $s(t)$ is $-x$, which is larger than $-N$.
- Similarly, if $T(t) = N$, slot machine B is selected regardless of the signal level $s(t)$ because $T(t) = N > s(t)$ always holds since the maximum of $s(t)$ is $+x$, which is smaller than $N$.
- If $-N + 1 \leq T(t) \leq N - 1$, the decision of selecting slot machine A or B depends on the signal level of $s(t)$.
    - If $s(t)$ is given by $+x$, the decision is to select machine A because $s(t) = +x$ is greater than $N - 1$.
    - Conversely, if $s(t)$ is given by $-x$, the decision is to select machine B because $s(t) = -x$ is smaller than $-N + 1$.

Furthermore, the incoming signal $s(t)$ contains inherent correlations, as discussed in Sections 1 and 2. Concerning the fact that $s(t)$ under study is a two-level signal train, we can think of the probability where the signal level $s(t + 1)$ at time $t + 1$ is different from $s(t)$ at time $t$; that is, $s(t + 1) = +x$ results after $s(t) = -x$ or $s(t + 1) = -x$ after $s(t) = +x$. Since the autocorrelation between two consecutive timings is given by $\lambda$, such a signal level changing probability is given by $\mu = (1 - \lambda)/2$. Conversely, the probability of exhibiting the same signal level is given by $1 - \mu = (1 + \lambda)/2$.

Therefore, such stochastic processes are represented by conditional probabilities given by
$$\Pr(s(t + 1) = \pm x \mid s(t) = \mp x) = \mu \text{ and } \Pr(s(t + 1) = \pm x \mid s(t) = \pm x) = 1 - \mu, \quad (4)$$
where Pr denotes probability. The important aspect is that the internal status of the decision-maker, represented by $T(t)$, is tightly coupled with the correlated time series subjected to the system as well as the betting results of the slot machine playing, which is specified by $P_A$ and $P_B$.

The behavior of the revision of *T(t)* is described by the following cases:
- If $T(t) = -N$, slot machine A is always selected. The threshold is updated as
$$T(t+1) = \begin{cases} T(t)+1 & \text{(when machine A fails with probability } 1-P_A) \\ T(t) & \text{(when machine A wins with probability } P_A) \end{cases}.$$
- If $T(t) = N$, slot machine B is always selected. The threshold is updated as
$$T(t+1) = \begin{cases} T(t) & \text{(when machine B wins with probability } P_B) \\ T(t)-1 & \text{(when machine B fails with probability } 1-P_B) \end{cases}.$$
- If $-N \leq T(t) \leq N-1$,
When slot machine A is selected, the threshold is updated as
$$T(t+1) = \begin{cases} T(t)+1 & \text{(when machine A fails with probability } 1-P_A) \\ T(t)-1 & \text{(when machine A wins with probability } P_A) \end{cases},$$
and when slot machine B is selected, the threshold is updated as
$$T(t+1) = \begin{cases} T(t)+1 & \text{(when machine B wins with prob. } P_B) \\ T(t)-1 & \text{(when machine B fails with prob. } 1-P_B) \end{cases}.$$
It should be noted that regardless of the machine selection and betting result, the threshold level always increases or decreases in this case, meaning that the same threshold level is not allowed.

The procedure summarized above is a special case of the principle shown in Section 2 by specifying the parameters therein by $k = \Delta = \Omega = \alpha = 1$. In addition, we have to emphasize that the upper and lower limits of *T(t)* is newly posed when the decrement or increment of the threshold is not permitted beyond the range between *–N* and *N*. Hereafter, we refer to this as the stopping rule. This setting is the simplest case for the laser chaos decision-maker. We use this simplest case to keep our analysis model from being too complicated. Cases with other settings may be possible by extending our proposed scheme, but this will be a future project.

To theoretically deal with the abovementioned seemingly complex situations, we introduce a set $\boldsymbol{v}_t = (T(t), s(t))$, which represents the state of the model at time *t*. The space spanned by $\boldsymbol{v}_t$ is $\{-N, -N+1, \cdots, N-1, N\} \times \{\pm x\}$.

Herein, we can characterize the state transition probability between two states. Let, for example, the current state is specified by (*i*, +*x*) while *T(t)* is not at the border; i.e., $-N+1 \leq i \leq N-1$. Here, we consider the probability of the state transition as (*i* + 1, –*x*). It should be noted that the decision is to select machine A in this given situation (*i*, +*x*) since the signal level +*x* is larger than the current threshold *T(t)*. In this state transition from (*i*, +*x*) to (*i* + 1, –*x*), the threshold is incremented ($i \to i + 1$) and the incoming signal level is reversed (+*x* → –*x*). Such a situation occurs when the slot machine A playing is unsuccessful and the incoming signal level is flipped, whose probability is given by $(1 - P_A)\mu$. Similarly, all transition probabilities are determined.

The notion of correlated random walk allows us to summarize such transitions in a unified manner [22,23]. We first introduce the probability of the state $\boldsymbol{v}_t$ by $\pi_t(\boldsymbol{v}) = \pi_t(i, \sigma)$, meaning the probability of the state with $T(t) = i$ and $s(t) = \sigma$. In addition, we define a probability vector $\boldsymbol{\pi}_t(i)$, which is given by
$$\boldsymbol{\pi}_t(i) = \begin{bmatrix} \pi_t(i, +x) \\ \pi_t(i, -x) \end{bmatrix}, \tag{5}$$
which combines the probabilities involving the threshold level being *i* for different signal levels of the time series (+*x* and –*x*).

We denote the probability of the threshold being $i$ at time $t$, regardless of the incoming signal level, by $\pi_t(i)$, which is mathematically equivalent to the $L^1$-norm of $\boldsymbol{\pi}_t(i)$. That is,

$$\pi_t(i) = \pi_t(i, +x) + \pi_t(i, -x) = |\pi_t(i, +x)|_1 + |\pi_t(i, -x)|_1 = \|\boldsymbol{\pi}_t(i)\|_1. \tag{6}$$

Based on these preparations, the recurrent formulae of $\boldsymbol{\pi}_t(i)$ leads us to precisely characterize the behavior of the system.

**[CASE 1]**

The probability vector for the case when the threshold is between $-N + 1$ and $N - 1$ at time $t + 1$ is given by

$$\boldsymbol{\pi}_{t+1}(i) = \boldsymbol{Q}(i-1)\boldsymbol{\pi}_t(i-1) + \boldsymbol{P}(i+1)\boldsymbol{\pi}_t(i+1), \tag{7}$$

where the matrices $\boldsymbol{P}$ and $\boldsymbol{Q}$ are given by

$$\boldsymbol{P}(i) = \begin{bmatrix} P_A(1-\mu) & (1-P_B)\mu \\ P_A\mu & (1-P_B)(1-\mu) \end{bmatrix} \tag{8}$$

$$\boldsymbol{Q}(i) = \begin{bmatrix} (1-P_A)(1-\mu) & P_B\mu \\ (1-P_A)\mu & P_B(1-\mu) \end{bmatrix}. \tag{9}$$

Equation (7) clearly implies that the probability vector of the threshold being $i$ comprises the transitions from the states with the thresholds being $i - 1$ and $i + 1$. The elements of the matrices $\boldsymbol{P}(i)$ and $\boldsymbol{Q}(i)$ are intuitively easily understood by the following. The dynamics given by equation (7) are schematically illustrated in Figure 4(a).

The matrix $\boldsymbol{P}(i)$ concerns the probability of *decrementing* the threshold level. For example, the (1,1)-element of $\boldsymbol{P}(i)$, or $\boldsymbol{P}_{1,1}(i)$, represents the probability of the transition from the state $(i, +x)$ to $(i - 1, +x)$. The state $(i, +x)$ indicates that the decision is to select machine A. The decrement of the threshold indicates that the result is a win. The probability of consecutive identical signal levels is given by $1 - \mu$. Hence, $\boldsymbol{P}_{1,1}(i) = P_A (1 - \mu)$. Similarly, $\boldsymbol{P}_{1,2}(i)$ means the probability of the transition from the state $(i, -x)$ to $(i - 1, +x)$; the difference is the change of the polarity of the incoming signal level. Therefore, $\boldsymbol{P}_{1,2}(i) = (1 - P_B) \mu$. Similarly, $\boldsymbol{P}_{2,1}(i)$ corresponds to the probability of the transition from the state $(i, +x)$ to $(i - 1, -x)$, and $\boldsymbol{P}_{2,2}(i)$ corresponds to the transition from $(i, -x)$ to $(i - 1, -x)$. The blue arrows in Figure 4(a) schematically represent the role of the matrix $\boldsymbol{P}(i)$, which concerns the decrementing of the threshold level.

Conversely, the matrix $\boldsymbol{Q}(i)$ concerns the probability of *incrementing* the threshold level. $\boldsymbol{Q}_{1,1}(i)$, for example, represents the probability of the transition from the state $(i, +x)$ to $(i + 1, +x)$, meaning that the threshold is incremented while the signal level is unchanged. This situation represents the decision to select machine A, the result is lost, and the polarity of the incoming signal is the same; the corresponding probability is given by $(1 - P_A)(1 - \mu)$. Similarly, other elements of $\boldsymbol{Q}(i)$ are specified in a straightforward manner. The red arrows in Figure 4(a) schematically represent the role of the matrix $\boldsymbol{Q}(i)$, which concerns the incrementing of the threshold level.

**[CASE 2]**

The probability vector for the case when the threshold is at the edge on the negative side, $-N$ at time $t + 1$ is specified by

$$\boldsymbol{\pi}_{t+1}(-N) = \boldsymbol{P}(-N)\boldsymbol{\pi}_t(-N) + \boldsymbol{P}(-N+1)\boldsymbol{\pi}_t(-N+1). \tag{10}$$

Edges are to be treated carefully in this case. First, $\boldsymbol{P}(-N + 1)$ in the second term on the right-hand side of equation (10) describes the transition of the decrement of the threshold level from $-N + 1$ to $N$, which has already been defined in equation (8). Second, since there are no threshold levels smaller than $-N$, the transitions involving increments or any $\boldsymbol{Q}$ matrix is not included in equation (10). Third, what is different from CASE 1 above is that the threshold level can be maintained at the edges, which is indicated by the first term on the right-hand side of equation (10). More specifically, the $\boldsymbol{P}$ matrix at $-N$ is given by

$$\boldsymbol{P}(-N) = \begin{bmatrix} P_A(1-\mu) & P_A\mu \\ P_A\mu & P_A(1-\mu) \end{bmatrix}. \tag{11}$$

$\boldsymbol{P}_{1,1}(-N)$ means the state transition from $(-N, +x)$ to $(-N, +x)$. This corresponds to the decision to select machine A, the result is a win, and the signal polarity is unchanged. Therefore $\boldsymbol{P}_{1,1}(-N) = P_A (1 - \mu)$. Similarly, $\boldsymbol{P}_{1,2}(-N)$ means the state transition from $(-N, -x)$ to $(-N, +x)$; what is different from $\boldsymbol{P}_{1,1}(-N)$ is the change in polarity. Hence, $\boldsymbol{P}_{1,2}(-N) = P_A \mu$. Likewise, $\boldsymbol{P}_{2,1}(-N)$ and $\boldsymbol{P}_{2,2}(-N)$ can be obtained. The blue arrows in Figure 4(b) illustrates the role of the matrix $\boldsymbol{P}(-N)$, which concerns keeping the same threshold level.

**[CASE 3]**
Similar to CASE 2, the probability vector for the case when the threshold is $N$ at time $t + 1$ is specified by

$$\boldsymbol{\pi}_{t+1}(N) = \boldsymbol{Q}(N-1)\boldsymbol{\pi}_t(N-1) + \boldsymbol{Q}(N)\boldsymbol{\pi}_t(N). \tag{12}$$

The meaning of equation (12) is similar to equation (10). $\boldsymbol{Q}(N-1)$ in the right-hand side of equation (12) has been already defined in equation (9). As in CASE 2, the threshold level can be maintained at the edge, which is shown by $\boldsymbol{Q}(N)$ in equation (12). This is given by

$$\boldsymbol{Q}(N) = \begin{bmatrix} P_B(1-\mu) & P_B\mu \\ P_B\mu & P_B(1-\mu) \end{bmatrix}. \tag{13}$$

$\boldsymbol{Q}_{1,1}(N)$ means the state transition from $(N, +x)$ to $(N, +x)$. This corresponds to the decision to select machine B, the result is a win, and the signal polarity is unchanged. Therefore $\boldsymbol{Q}_{1,1}(N) = P_B (1 - \mu)$. Similarly, $\boldsymbol{Q}_{1,2}(N)$ indicates the state transition from $(N, -x)$ to $(N, +x)$; what is different from $\boldsymbol{Q}_{1,1}(N)$ is the change in polarity. Hence, $\boldsymbol{Q}_{1,2}(N) = P_B \mu$. Likewise, $\boldsymbol{Q}_{2,1}(N)$ and $\boldsymbol{Q}_{2,2}(N)$ can be obtained. The red arrows in Figure 4(c) illustrate the role of the matrix $\boldsymbol{Q}(N)$, which concerns keeping the same threshold level.

Finally, a remark is needed for the matrix $\boldsymbol{P}$ at $N$ and matrix $\boldsymbol{Q}$ at $-N$, which should be different from the one given by equations (8) and (9), and are given by

$$\boldsymbol{P}(N) = \begin{bmatrix} (1-P_B)(1-\mu) & (1-P_B)\mu \\ (1-P_B)\mu & (1-P_B)(1-\mu) \end{bmatrix}, \tag{14}$$

$$\boldsymbol{Q}(-N) = \begin{bmatrix} (1-P_A)(1-\mu) & (1-P_A)\mu \\ (1-P_A)\mu & (1-P_A)(1-\mu) \end{bmatrix}. \tag{15}$$

This is because the decision at the edges does not depend on the incoming signal level. For example, with the threshold at $N$, the decision is always to select machine B because both signal level $+x$ and $-x$ is smaller than the threshold. Hence $\boldsymbol{P}_{1,1}(N)$ means the probability of the state transition from $(N, +x)$ to $(N - 1, +x)$, meaning that the decision is to select machine B, the result is a loss, and the polarity of the signal is unchanged. Therefore $\boldsymbol{P}_{1,1}(N) = (1 - P_B) (1 - \mu)$. Similarly, all other elements in equations (14) and (15) are specified. The blue arrows in Figure 4(c) and the red arrows in Figure 4(b) illustrate $\boldsymbol{P}(N)$ and $\boldsymbol{Q}(-N)$, respectively.

Figure 5 summarizes the chains of the probability vector $\boldsymbol{\pi}_t(i)$ by equations (7), (10), and (12). The blue arrows, which regard the decrement of the threshold level, are induced by either a win by selecting machine A or loss by selecting machine B. In contrast, the red arrows, which represent the increment of the threshold level, are triggered by either a win by selecting machine B or loss by selecting machine A. The thresholds at the edge ($-N$ and $N$) involve arrows of transitions to an identical threshold.

Finally, the CDR can be discussed using the probabilities defined above. Assume that the correct decision is to select machine A. The selection of machine A is realized excessively in the following two cases:
1. The threshold is $-N$. In this case, both signal levels $-x$ and $+x$ result in the decision to choose machine A.
2. When the threshold is between $-N + 1$ and $N - 1$, the input signal level of $+x$ results in the decision to choose machine A.

Hence, the probability of selecting machine A at time $t$, denoted by $CDR^{(theory)}(t)$, is given by

$$CDR^{(theory)}(t) = \pi_t(-N, -x) + \pi_t(-N, +x) + \sum_{i=-N+1}^{N-1} \pi_t(i, +x). \quad (16)$$

## 5. Evaluation

With the theoretical model shown in Section 4, we can calculate the time evolution of the probability vector $\boldsymbol{\pi}_t(i)$ and its $L^1$-norm $\pi_t(i)$ from any initial conditions. Consequently, $CDR^{(theory)}(t)$ is derived by equation (16).

Here, we examine the case when the reward probabilities are given by $P_A = 0.9$ and $P_B = 0.7$ and assume that $N$ is given by 2, meaning that the number of threshold levels is 5. Herein, the initial probability vector is given by $\boldsymbol{\pi}_1(0) = (0.5, 0.5)$ while assuming all the other vectors are zero. The autocorrelation coefficient $\lambda$ specifies the time-correlated, two-level signal trains.

Figure 5(b) shows the analytically calculated chains of probability vectors. As time evolves, the probability vector at the edge ($i = -2$) increases, indicating a high likelihood of choosing machine A, which is the correct decision (since $P_A > P_B$).

To examine the mechanism more deeply, Figures 6(a), 6(b), and 6(c) demonstrate the time evolution of the probability when the threshold is at level $i$ ($i = -2, -1, 0, 1, 2$) when the autocorrelation $\lambda$ is specified by $-0.8$, 0, and 0.8, respectively. What is commonly observed in these figures is that $\pi_t(-2)$, indicated by blue curves, increase as the time elapses, leading to a high chance of selecting machine A or correct decision-making. Meanwhile, $\pi_t(2)$, indicated by green curves, exhibit approximately 0.2 at a time step of 25 when $\lambda$ is 0.8 (Figure 6(c)), whereas it shows nearly zero at the same timing when $\lambda$ is $-0.8$ (Figure 6(a)). This indicates that the probability of choosing machine B, which is the wrong decision, is not negligible when $\lambda = 0.8$.

From another perspective, the blue, red, and yellow markers in Figure 6(d) characterize the probabilities of the threshold at $t = 1000$, which is written as $\pi_{1000}(i)$, when the autocorrelation is specified for $\lambda$ values given by $-0.8$, 0, and 0.8 respectively. We can clearly observe a large probability greater than 0.6 about the threshold level of $-2$, regardless of $\lambda$ values.

It is remarkable that for $\lambda = -0.8$, the probability monotonically decreases as the threshold increases, whereas for $\lambda = 0.8$, the probability increases when the threshold increases from 0 to 2. Even with zero autocorrelation ($\lambda = 0$), a slight increase in probability is observed at the threshold level of 2. We assume that a positive autocorrelation tends to conduct similar decisions consecutively, and hence the decision can be locked in a status, which is actually not the optimal one. Indeed, a related tendency is observed in Figures 6(a), 6(b), and 6(c), where the dynamic change of probabilities, most notably by $\pi_t(0)$ indicated by orange curves, exhibit a strong oscillatory behavior with $\lambda = -0.8$, whereas it is attenuated when $\lambda = 0.8$.

As discussed in Section 4, the decision-making ability can be theoretically derived as $CDR^{(theory)}(t)$, given in equation (16) using the probability model. We examined $CDR^{(theory)}(t)$ depending on a variety of conditions. Herein, the reward probabilities ($P_A$, $P_B$) and the number of threshold levels specified by $N$ are summarized in Table 1, which are the same as discussed in Section 3 and Figure 2. For example, Figure 7(a) concerns the case ($P_A$, $P_B$) = (0.9, 0.3) and $N$ = 2. The red curves in Figure 7 show $CDR^{(theory)}(1000)$ as a function of autocorrelation coefficient $\lambda$ ranging from –0.95 to 0.9 with 0.05 interval. In addition, $\lambda$ = –0.99 is examined. For all cases in Figure 7, the maximum $CDR^{(theory)}(1000)$ are obtained when the autocorrelation coefficient is negative, indicated by red arrows therein, which coincide with the numerical observations shown in Figure 2.

Furthermore, we numerically simulate the correct decision rate $CDR(t)$ defined in equation (3) based on the original decision-making algorithm described in Section 3 while adapting the stopping rule in Section 4. The results are shown by the blue curves in Figure 7. We observe in all panels in Figure 7 that the results from theory (red) and simulation (blue) match well with each other. Additionally, while the blue marks exhibit fluctuations since they are obtained as a statistical average via numerical results, the results in red marks are smooth because they are analytically derived based on the theory described in Section 4.

## 6. Conclusion

In this study, we construct a theoretical model to account for the acceleration of decision-making by correlated time sequences. Previous studies have shown that the solution to the two-armed bandit problem is accelerated by negative autocorrelation inherent in the time series subjected to the decision-making system. However, its underlying mechanisms are unclear. We begin the discussion by clarifying the impact of time-domain correlation on decision-making by utilizing time series with specific autocorrelation designed via Fourier transform surrogate. Coinciding with the prior reports of using experimentally observed laser chaos time series, we confirm that the negative autocorrelation accomplishes superior decision-making performance. The difficulties in understanding the underlying mechanism of such acceleration stem from the fact that multiple entities are involved: the dynamical reconfiguration of the internal status of the decision-maker (the threshold level and its revision), time-domain structure of the incoming time series, and stochastic attributes of the environment (reward probability of slot machines). The theoretical model of this study unifies these entities based on correlated random walks. Furthermore, the decision-making performance obtained analytically by the theoretical model agrees with the numerical results from simulations, which validates the proposed theory. Additionally, this indicates that the optimal autocorrelation for maximizing can be obtained through the model without executing enormous numerical simulations. The proposed scheme to select the best laser chaos with the best autocorrelation can accelerate performance in applications such as wireless communication systems [11–13]. This study constitutes a foundation of the intellectual mechanism enhanced by correlated time series, which is important for future information and communications technology.

The laser chaos decision-maker can quickly solve MAB problems with Giga Hz order decisions. Therefore, it will be possible to optimize decisions in wireless communication systems in real-time. However, a dedicated device for the laser chaos decision-maker is necessary. In the meantime, a chip-scale photonic implementation has been recently demonstrated [25] on the basis of the recent advancements of integrated photonics technologies, indicating the potential of system integration and miniaturization.

## Acknowledgments

This work was supported in part by the CREST project (JPMJCR17N2) funded by the Japan Science and Technology Agency and Grants-in-Aid for Scientific Research (JP20H00233) funded by the Japan Society for the Promotion of Science.

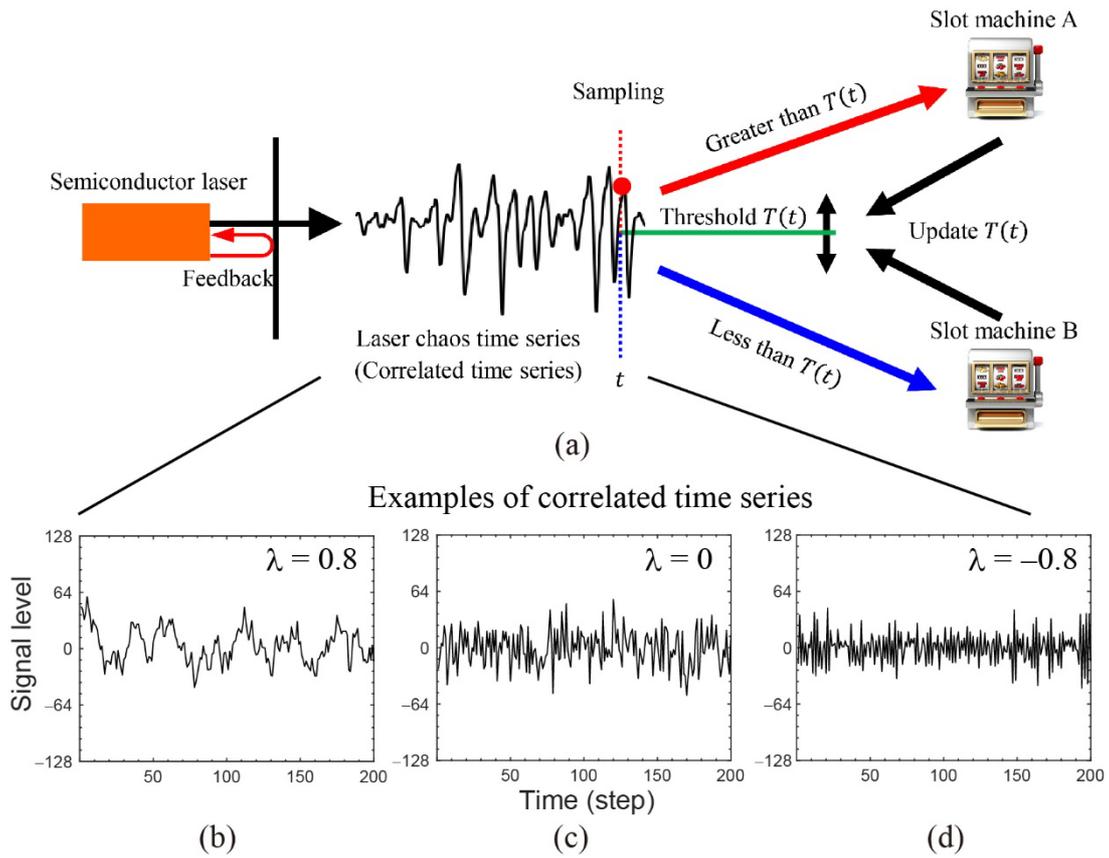

Figure 1: (a) Decision-making architecture using laser chaos time series or correlated time series. (b, c, d) Snapshots of correlated time series generated via Fourier transform surrogate. The correlation is specified by the parameter $\lambda$. (b) $\lambda = 0.8$ (strong positive autocorrelation), (c) $\lambda = 0$ (zero correlation), and (d) $\lambda = -0.8$ (strong negative autocorrelation).

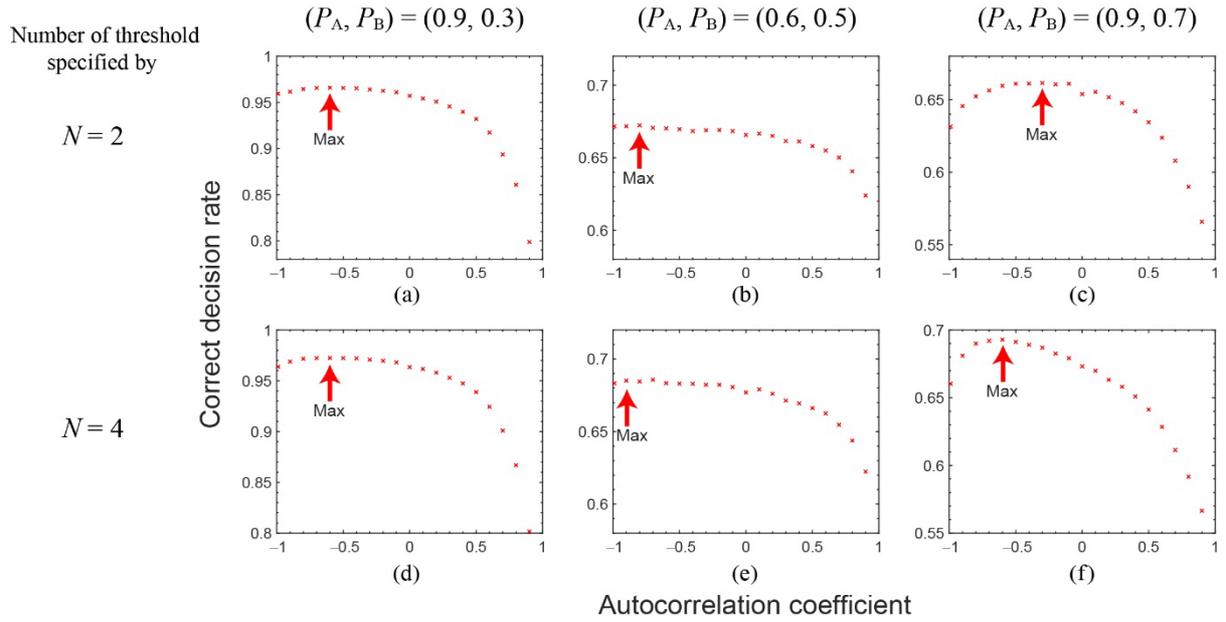

Figure 2: Correct decision rate at *t* = 1000 as a function of the autocorrelation coefficient $\lambda$ in various reward environments and a different number of threshold levels. It is to be noted that the maximum performance is obtained all by negatively correlated time series. (a) $P_A = 0.9, P_B = 0.3, N = 2$, (b) $P_A = 0.6, P_B = 0.5, N = 2$, (c) $P_A = 0.9, P_B = 0.7, N = 2$, (d) $P_A = 0.9, P_B = 0.3, N = 4$, € $P_A = 0.6, P_B = 0.5, N = 4$, and (f) $P_A = 0.9, P_B = 0.7, N = 4$.

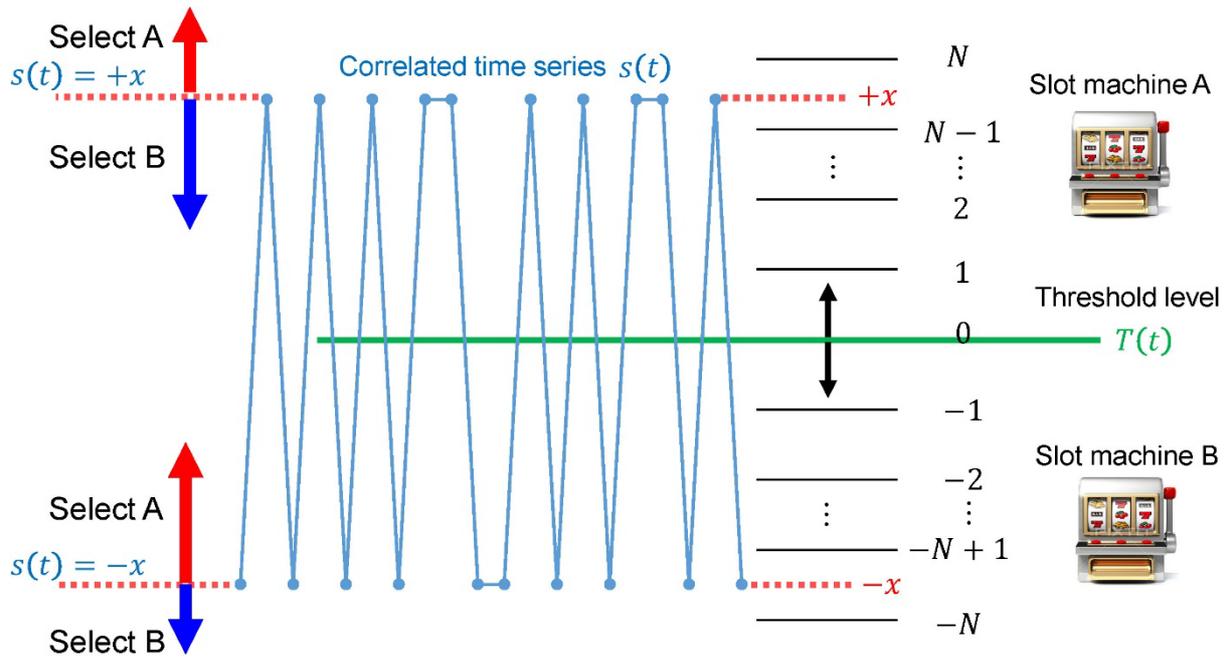

Figure 3: The theoretical model of the decision-making based on correlated time series and reconfiguration of the threshold.

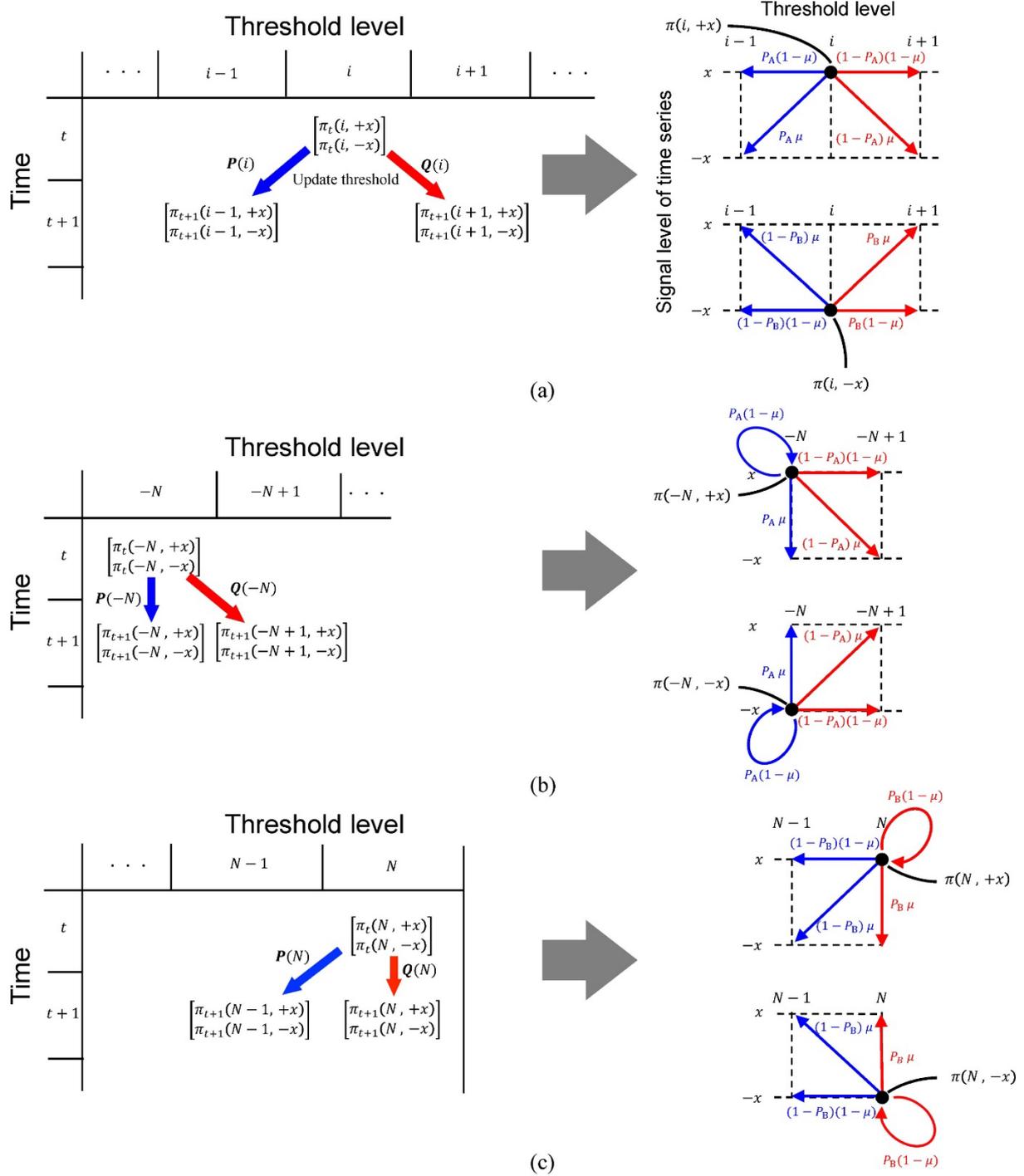

Figure 4: Transitions probability vectors of the model. (a) The matrices of **P** and **Q** concern the increment and decrement of the threshold level. (b, c) When the threshold is at the edge ($-N$ or $N$), transitions to an identical threshold should be considered, which are represented by **P**($-N$) and **Q**($N$) in (b) and (c), respectively.

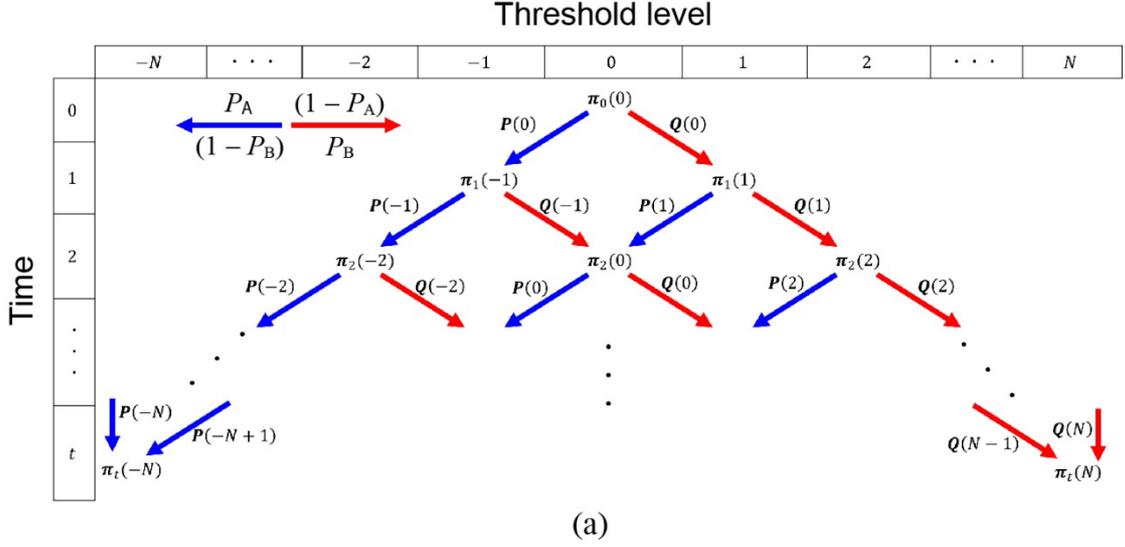

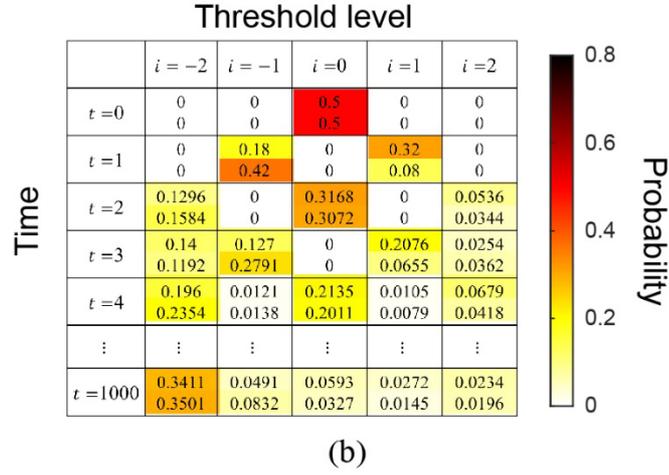

Figure 5: (a) Chains of the probability vector $\pi_t(i)$ given by equations (7), (10), and (12). (b) An example of the evolution of probability vector $\pi_t(i)$ when the initial condition is $\pi_0(0) = (0.5, 0.5)$, the autocorrelation coefficient $\lambda$ is $-0.8$, the threshold number is specified by $N = 2$, and the reward environment is $(P_A, P_B) = (0.9, 0.7)$.

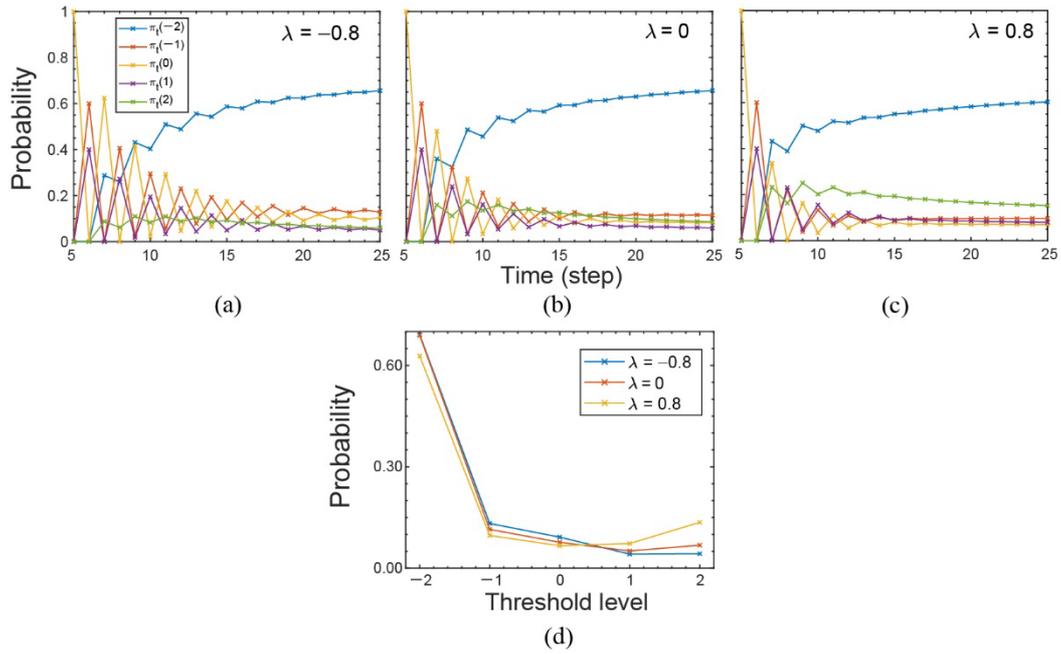

Figure 6: (a,b,c) The time evolution of the probabilities that occupy each of the threshold levels. The reward probabilities are given by $(P_A, P_B) = (0.9, 0.7)$. The autocorrelation coefficient $\lambda$ is configured differently: (a) $\lambda = -0.8$, (b) $\lambda = 0$, and (c) $\lambda = 0.8$. (d) Comparison of the probabilities occupying threshold level with different autocorrelation coefficients at $t = 1000$.

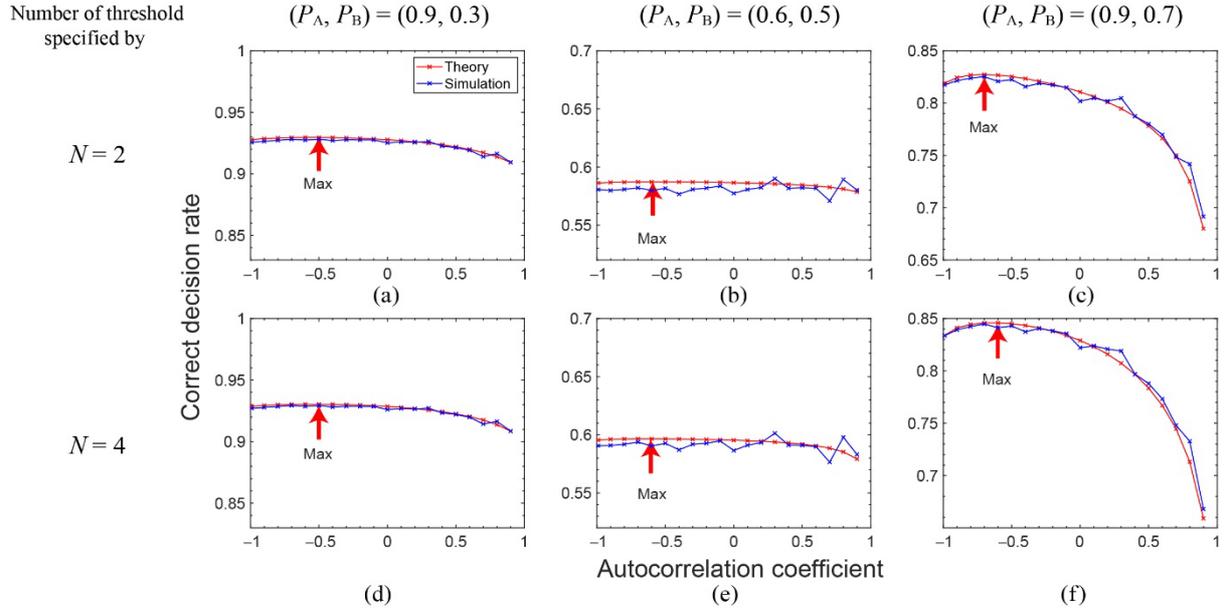

Figure 7: Correct decision rate at $t = 1000$ derived by the proposed theoretical model (red) and numerical simulations (blue) as a function of the autocorrelation coefficient $\lambda$. The reward probability settings and number of threshold levels are the same as in Figure 2. The threshold level is specified by $N = 2$ in (a), (b), and (c), whereas $N = 4$ in (d), (e), and (f). In (a) and (d), the reward probability is given by $(P_A, P_B) = (0.9, 0.3)$. In (b) and (e), $(P_A, P_B) = (0.6, 0.5)$. In (c) and (f), $(P_A, P_B) = (0.9, 0.7)$.

Table 1: The settings of the reward probabilities of slot machines ($P_A$ and $P_B$) and the parameter $N$ that specifies the number of threshold levels ($2N + 1$).

| Figure | $P_A$ | $P_B$ | $N$ |
|---|---|---|---|
| 2(a), 7(a) | 0.9 | 0.3 | 2 |
| 2(b), 7(b) | 0.6 | 0.5 | 2 |
| 2(c), 7(c) | 0.9 | 0.7 | 2 |
| 2(d), 7(d) | 0.9 | 0.3 | 4 |
| 2(e), 7(e) | 0.6 | 0.5 | 4 |
| 2(f), 7(f) | 0.9 | 0.7 | 4 |